# Two-stage Image Classification Supervised by a Single Teacher Single Student Model

Jianhang Zhou[†], Shaoning Zeng[†]
mb85405@um.edu.mo,
zsn@outlook.com
Bob Zhang[*]
bobzhang@um.edu.mo

PAMI Research Group
Department of Computer and
Information Science
University of Macau
([†] Both authors contributed equally)

**Abstract**

The two-stage strategy has been widely used in image classification. However, these methods barely take the classification criteria of the first stage into consideration in the second prediction stage. In this paper, we propose a novel two-stage representation method (TSR), and convert it to a Single-Teacher Single-Student (STSS) problem in our two-stage image classification framework. We seek the nearest neighbours of the test sample to choose candidate target classes. Meanwhile, the first stage classifier is formulated as the teacher, which holds the classification scores. The samples of the candidate classes are utilized to learn a student classifier based on L2-minimization in the second stage. The student will be supervised by the teacher classifier, which approves the student only if it obtains a higher score. In actuality, the proposed framework generates a stronger classifier by staging two weaker classifiers in a novel way. The experiments conducted on several face and object databases show that our proposed framework is effective and outperforms multiple popular classification methods.

## Introduction

Image classification is one of the most crucial techniques in computer vision. While one-step classification might not be credibly adequate, two-stage image classification has been successful in many tasks, i.e., face recognition [1-3] and object recognition [4]. Real-world recognition tasks often contain a lot of complicated data and conditions. For this reason, the discriminative ability of one single classifier is likely to fail in picking the best result. Thus, it is reasonable to use two-stage classification to perform coarse and fine classification. Since the final result is determined by a subset of the classes, the complexity of the distribution in data is reduced [4]. In addition, according to the probability estimation, it is more effortless to choose multiple candidate classes containing the right class than to find out if one single class is the right class. For instance, the Top-1 accuracy of ImageNet is hardly equal to the Top-5 accuracy in any condition. Therefore, two-stage classification, in various implementations, is more promising than others.

Two-stage methods have been popular for a long time [5-7]. For example, Xu et al. proposed a two-phase test sample representation (TPTSR) method [5], which used all training samples to represent a test sample in order to exploit its nearest neighbours in the first stage, before organizing these nearest neighbours to represent the test sample. The WSRC (weighted sparse representation for classification) [2] exploited weights of the representation to seek a sparser representation, which is a form of the sparse representation method implemented as a two-stage method. The RAMUSA [8] algorithm performed multi-task learning by using a multi-stage method, which is similar to the idea of two-stage methods. In addition, the idea of two-stage classification is helpful when implemented to other problems, such as coarse-to-fine frameworks [9-10], and face recognition [11]. From the above discussion, we believe that all of these two stage classification methods only paid attention to the two classifiers, without considering the relationship of the classification criteria, or scores, between the two stages.

In contrast to this, a Teacher-Student model has a much clearer role of definitions for the two classifiers. Recently, You et al. proposed g-SVM for solving the single-teacher multi-students problem [12]. The multi-teacher single-student problem was solved by a multi-teacher networks model [13]. Zheng et al. combined GAN with the teacher-student problem and achieved effective results [14]. In our opinion, the teacher-student model is a special case of the two-stage classification. However, the teacher classifier cannot reduce the computation load of students, which is different from the first classifier in conventional two-stage methods. What is more, no such findings are available to solve the single-teacher single-student problem by using a score-based prediction mechanism.

Both the Two-Stage and Teacher-Student classification methods have various implementations. Among them, linear methods show promising performances, i.e., Sparse Representation (SR) and SVM. SVM proposed in 1995 [15] is a powerful classifier in different classification tasks [16-17]. Currently, there are several variants and applications. The sparse representation classifier (SRC) [18] shows effective performances and robustness in image classification as well by taking advantage of the role of sparsity. Another powerful representation method is the collaborative representation classifier (CRC) [19]. By putting emphasis on collaborative representation, CRC improves the efficiency of SRC. There are fusion works of SRC and CRC [20-21], trying to keep a balance between sparsity and collaborative representation. Nevertheless, no such work can fuse them in a two-stage classification framework.

In this paper, we propose a novel framework for image classification using a two-stage representation method and formulate the two-stage classification problem to a single-teacher single-student problem. We name it **T**wo-**S**tage image classification supervised by a **S**ingle **T**eacher **S**ingle **S**tudent model (TS-STSS) and utilize sparse representation to implement the algorithm. In the first stage of classification, the teacher classifier makes classification and seeks the nearest classes to the test sample, which is denoted as 'candidate classes'. Then, a 'candidate set' containing the training samples can be organized according to 'candidate classes'. In the second stage, we represent the test sample and perform classification using the candidate set. Next, we use the single-teacher single-student model to make a decision based on scores generated from the teacher and student classifier. Generally speaking, our proposed two-stage representation method is supervised by a teacher classifier in image classification. The contributions of our work can be summarized in four aspects as follows:

1) We propose a novel two-stage representation method for image classification.

2) We formulate the decision-making problem between results of both stages to a single-teacher single-student problem, and solve it using a score-based mechanism.

3) We implement TS-STSS via L1-minimization (sparse representation) and L2-minimization (collaborative representation).

The remainder of this paper will be organized as follows. In Section 2, we first describe the two-stage test representation method and the single-teacher single-student strategy, before proposing our classification framework. In Section 3, experimental and comparison results on image datasets will be demonstrated to show the effectiveness and performance of our proposed method. Section 4 concludes this paper.

## 2. The Method

Our proposed method TS-STSS is a novel Two-Stage classification supervised by the Single-Teacher Single-Student model, where the implementation is based on sparse supervised representation [22]. In the first stage, a sparse representation-based classifier via L1-minimization is learned as the teacher classifier, which computes the distances (or scores) to select a set of candidate classes. Then, in the second stage, one single student classifier based on the faster L2-minization, is trained using the samples of all candidate classes. Meanwhile, the scores of each class are generated. With the supervision via scoring of the teacher classifier, the student classifier in the second stage is capable of generating the final result. The detailed process of TS-STSS is depicted in the following sub-sections.

### 2.1 Two-Stage (TS) Representation

The representation procedure is performed in two stages. Specifically, in the first stage, all training samples are used to represent the test sample in a linear combination:

$$y = \theta_1 x_1 + \theta_2 x_2 + ... + \theta_m x_m \quad (1)$$

where $y$ is the test sample, and the $\theta_i$ is the coefficient of the $i$ th instance in the linear combination, $x_i \in \Re^{r \times 1}$ is the column vector of the $i$ th instance, and $m$ is the number of instances in training set. For each class, we calculate its deviation with test sample by:

$$V_j = \left\| y - \sum_{i=1}^{m} \theta_{j,i} x_{j,i} \right\|^2 \quad (2)$$

where $V_j$ denotes the deviation of the $j$ th class, $\theta_{j,i}$ and $x_{j,i}$ are the $i$ th coefficient and $i$ th sample of the $j$ th class.

According to equation (2), we pick $N$ nearest neighbors and append their corresponding class label to the candidate classes set $C$. We denote a sample set gathering samples from $C$ as 'candidate set' $G$.

In the second stage, each test sample will be represented by samples in $G$ using a linear combination:

$$y = \hat{\theta}_1 x_1 + \hat{\theta}_2 x_2 + ... + \hat{\theta}_n x_n \quad (3)$$

Where $n$ denotes number of instances in $G$, and the $\hat{\theta}_i$ is the coefficient of $i$ th instance in the linear combination.

### 2.2 Single-Teacher Single-Student (STSS) Model

To take results from the first stage and second stage into consideration when classifying, we design a single-teacher single-student model, and solve it using a score-based mechanism. We define the classifier in the first stage as the teacher classifier, and classifier in the second stage as the student classifier. Then, we calculate the 'gate value' of the teacher and student classifiers respectively in order to make a comparison to decide the final result. In our strategy, we take the highest value of the teacher classifier and student classifier as the 'gate value'. We denote this solution as a single-teacher single-student model (STSS).

Firstly, we utilize a strong multi-class classifier $T$ as a teacher classifier. Then, we apply a faster classifier as the student classifier $ST$. Next, we use the teacher classifier $T$ to perform multi-class classification and obtain a score vector $S \in \Re^{K \times 1}$ for each class:

$$S_j = \delta(X_j) \tag{4}$$

where $S_j$ is the $j$ th instance of a score vector, $X_j$ denotes the training set of the $j$ th class, and $\delta(\cdot)$ is a score evaluation function to evaluate the score of the sample vector. Classes with the highest score will be selected as the classification result by the teacher classifier, and its corresponding score will be taken as the gate value $g$:

$$g = \max(S) \tag{5}$$

The gate value $g$ can also be regarded as the confidence of teacher ($T$). When the classification results of the student and the teacher are different, the final decision should be made between them. In this scenario, if the student ($ST$'s) learned highest score is higher than the gate value $g$, the $ST$'s classification result will determine the final result. Otherwise, the classification result of $T$ will be taken as the final result. The final result $z$ is determined as follows:

$$z = \begin{cases} L(S_{ST}), & \text{if } S_{ST} > S_T \\ L(S_T), & \text{otherwise} \end{cases} \tag{6}$$

where $L(\cdot)$ denotes the function mapping of a score to its corresponding class label, and $S_{ST}$, $S_T$ denote highest scores learned by a student $ST$ and a teacher $T$, respectively.

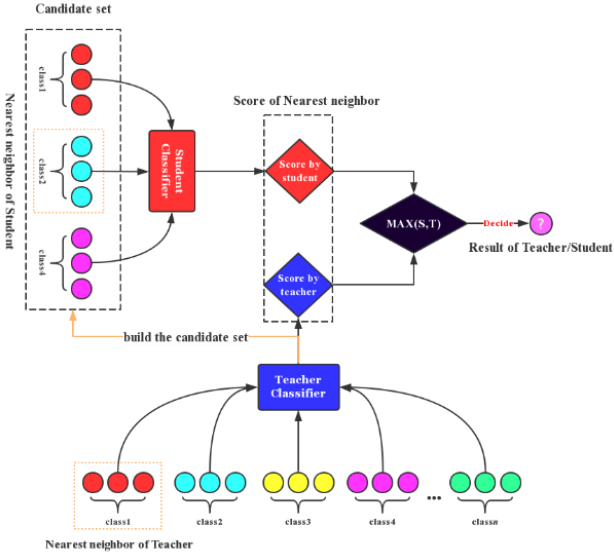

Figure 1: Two-stage image classification supervised by a teacher classifier.

It is obvious that the single-teacher single-student strategy is a two-fold learning strategy, which motivates us to combine it with a two-stage representation method, as both of the two learning strategies require two steps to perform classification.

## 2.3 Supervision of Teacher Classifier

Based on the two-stage representation method (TSR) and the single-teacher single-student model (STSS), we propose our two-stage image classification framework, named two-stage image classification supervised by a single-teacher single-student model (TS-STSS) for image classification. Figure 1 depicts the general idea of our method intuitively. First of all, in the initial stage, we apply the sparse representation classifier (SRC) [23] as teacher classifier using all training samples for representation:

$$\hat{\theta} = \arg\min_{\theta} \|\theta\|_1 \text{ s.t. } \|y - X\theta\|_2 < \varepsilon \tag{7}$$

where $\theta$ is the coefficient vector in linear combination, and $\varepsilon$ is the noise in $y$.

The candidate class set $C = \{C_1, C_2, ..., C_m\}$ is built by appending the class associated with the $M$ lowest deviations, and candidate set $G = \{X_{C_1}, X_{C_2}, X_{C_3}, ... X_{C_m}\}$. Next, the gate value can be described as follows:

$$\begin{aligned} g_j &= \frac{1}{n}\sum_{i=1}^{n} \|\tilde{y} - \Lambda_i X_i\|_2 - \|\tilde{y} - \Lambda_j X_j\|_2 \\ g^* &= \max(g) \end{aligned} \tag{8}$$

where $\Lambda_i = [0,...,0, \lambda_{i,1}, \lambda_{i,2},...,\lambda_{i,k}, 0,...,0]$, $X_i$ denotes training set of the $i$ th class, and $n$ represents the number of all classes.

Following this, in the second stage, we apply the collaborative representation classifier (CRC) whose classification speed is higher than SRC [19] as the student classifier using samples from the candidate set for representation:

$$\hat{\lambda} = \arg\min_{\lambda} \|y - \tilde{X}\lambda\|_2 \text{ s.t. } \|\lambda\|_q \leq \varepsilon \tag{9}$$

where $\lambda$ is the coefficient vector in linear combination, $\varepsilon$ is the noise in $y$, and $q$ can be 1 or 0. The solution of equation (9) is $\hat{\lambda} = (\hat{X} \cdot X + \lambda \cdot I)^{-1} X^T$.

The score learned by the student can be described as follows:

$$\begin{aligned} s_j &= \frac{1}{k}\sum_{i=1}^{k} \|y - \sigma_i \tilde{X}_i\|_2 - \|y - \sigma_j \tilde{X}_j\|_2 \\ s^* &= \max(s_j) \end{aligned} \tag{10}$$

where $\tilde{X}_i$ is the $i$ th instance in $G$, and $k$ is the number of instances in $G$.

If the classification results of the student and teacher are different from each other, we compare the score of the class identified by the student and teacher respectively before utilizing the decision-making method discussed in section 3.2:

$$\hat{z} = \begin{cases} L(s^*), & \text{if } s^* > g^* \\ L(g^*), & \text{otherwise} \end{cases} \quad (11)$$

where $L(\cdot)$ denotes the function mapping score $s$ and the gate value $g$ to its corresponding class label.

We summarize our proposed TS-STSS framework in Algorithm 1:

---
**Algorithm 1** TS-STSS classification framework
---
**Input**: Training set $X$, test sample $y$
**Output**: identity $I$
1: In the first stage, use SRC as the teacher classifier according to (7) to obtain a candidate set $C$ and classification result $R_t$.
2: According to (8), calculate the gate value $g$.
3: Perform the second phase classification using CRC as the student according to (9) and obtain the result of student $R_s$.
4: Calculate score $s$ learned by the student according to (10).
5:   **if** $R_t \neq R_s$ **then**

$$I = \begin{cases} L(s), & \text{if } s > g \\ L(g), & \text{otherwise} \end{cases}.$$

6:   **else**
7:     $I = R_t$.
8:   **end if**
10: **return** $I$

---

## 3. Experiments

To verify the effectiveness of our proposed method in different classification tasks, we performed several experiments on five datasets. Specifically, we tested our method on the FEI, MUCT, and YouTube facial datasets, the COIL-100 object dataset and the MNIST handwriting datasets, respectively. In addition, we conducted bench-mark experiments to other popular classifiers. The recognition rate was evaluated using a hand-out method, and we set different configurations for the different datasets. On COIL-100, MUCT and FEI, we increased the number of training samples in each class for every iteration and took the remaining samples of each class as test sample. Then, we calculated the average accuracy and maximum accuracy respectively. On MNIST and YouTube, we directly used the pre-divided training set and testing set. The experiments were executed using MATLAB R2018b on a PC with one 3.40GHz CPU and 16.0 GB RAMs.

## 3.1 Dataset description

As shown in Figure 2, we used five image datasets in total to evaluate our proposed method, including COIL-100 [24], MNIST handwriting digits [25], MUCT [26], FEI [27] and YouTubeFace [28], respectively. The details of each database are summarized in Table 1.

There are three facial datasets used in the experiments. The MUCT face database contains 3,755 face images from 276 people. The resolution of each image is 640*480 pixels. All images were captured by a CCD camera and stored in 24-bit RGB format. The FEI dataset is a face dataset containing 2,800 images from 200 people (14 images per person). The resolution of each original image is 640*480 pixels. In our experiments, we used the 24*96 pixels version. FEI and MUCT are relatively small, therefore, we wish to demonstrate our proposed TS-STSS works well in small datasets.

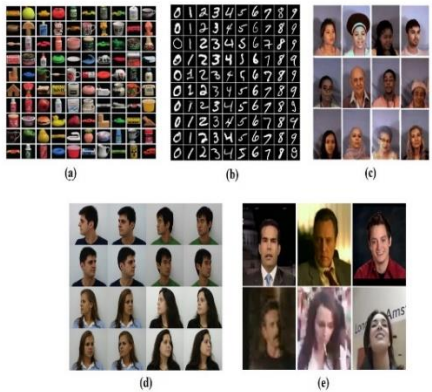

Figure 2: Image Datasets: (a) COIL-100, (b) MNIST,
(c) MUCT, (d) FEI, (e) YouTubeFace.

The last face dataset is YouTubeFace, which is a large-scale dataset. It is designed for studying unconstrained face recognition problems in video. In this dataset, 3,425 videos from 1,595 different people were collected from the YouTube website and labeled according to the LFW image collection method [29]. The resolution of each image is 32*32 pixels. In our experiments, we chose 1,283 classes with over 100 samples, and randomly selected 100 samples per class, 128,300 samples in total. Hence, we can evaluate the performance of TS-STSS for large-scale recognition.

The COIL-100 dataset (Columbia Object Image Library) is the object dataset, which collected 100 objects and contains 7,200 images in total with a black background. Each object has 72 images captured in different degrees by a CCD color camera. The resolution of each image is 32*32 pixels.

The MNIST hand-writing digits database is a hand-writing digits database built by LeCun et.al [25], containing 60,000 examples for a training set and 10,000 examples for a test set. The resolution of each image is 20*20 pixels. Each image was acquired from the center of 28*28 pixels from the original image and processed by a normalization algorithm.

| Database | Classes | Samples | Size | Dimension |
|---|---|---|---|---|
| FEI | 200 | 2,800 | 24*96 | 2-D |
| MUCT | 276 | 3,755 | 640*480 | 2-D |
| YouTubeFace | 1,595 | 620,951 | 32*32 | 2-D |
| COIL-100 | 100 | 7,200 | 32*32 | 2-D |
| MNIST | 10 | 7,0000 | 28*28 | 2-D |

Table 1: Configurations of the image databases in the experiments.

## 3.2 Face Recognition

We used the MUCT, FEI and YouTubeFace datasets to perform experiments on face recognition. From the results shown in Table 2, we notice that TS-STSS achieves the highest accuracy on the MUCT dataset, which is 92.06%, indicating its effectiveness on face recognition. Besides this, TS-STSS outperforms most of the classifiers in face recognition both in maximum and average accuracy (MUCT: 92.06%, 80.03%; FEI: 90.5% and 62.45%). As for the YouTubeFace dataset, the proposed method achieved a recognition rate of 90.94%, which is only 0.1% off the best result from [12]. Figures 3(a) and 3(b) show the accuracies generated by SRC, CRC and TS-STSS on the MUCT and FEI datasets respectively. In Figure 3(a), it is obvious that TS-STSS produced a better recognition performance no matter the training samples. In Figure 3(b), the accuracy of TS-STSS and SRC is quite competitive in beginning, while TS-STSS surpasses SRC after using seven training samples.

| Methods | Object | | Handwriting | Face | | | | |
|---|---|---|---|---|---|---|---|---|
| | COIL-100 | | MNIST | MUCT | | FEI | | YouTubeFace |
| | MAX | AVG | ACC | MAX | AVG | MAX | AVG | ACC |
| SRC | 76.97 | 73.77 | 95.96 | 90.15 | 79.01 | 89.80 | 57.95 | 84.91 |
| CRC | 70.84 | 65.23 | 82.83 | 85.83 | 76.87 | 74.75 | 49.26 | 72.12 |
| K-SVD | 61.58 | 58.62 | 82.87 | 77.09 | 70.26 | 65.88 | 40.95 | 53.26 |
| SVM | 58.94 | 53.54 | **98.60** | 28.44 | 24.78 | 57.13 | 40.95 | 78.90 |
| KNN | 74.56 | 70.02 | 95.00 | 67.94 | 57.97 | 69.63 | 48.55 | 90.00 |
| TPSTR[5] | 76.89 | 72.41 | 87.27 | 88.54 | 66.48 | 89.17 | 61.88 | 78.04 |
| STMS[12] | 77.19 | 72.43 | 95.59 | 89.64 | 75.26 | 89.66 | 61.32 | **91.04** |
| TS-STSS | **78.84** | **75.03** | 96.19 | **92.06** | **80.03** | **90.50** | **62.45** | 90.94 |

Table 2: Recognition rate comparisons to popular classifiers.

(Note: Unit of data is %, and bold figures indicate the best results.)

## 3.3 Object Recognition

We used the COIL-100 dataset to perform experiments on object recognition. As can be seen in Table 2, the highest accuracy achieved by TS-STSS is 78.84%, which is higher than SRC (76.97%) and CRC (70.84%), respectively. Noticeably, the highest improvement for average accuracy compared with SRC and CRC is 1.26% and 9.8% correspondingly, showing that the proposed method has a significant effect on object recognition. Figure 3(c) shows the accuracy generated by SRC, CRC and TS-STSS. We can observe directly that the gap between TS-STSS and SRC, CRC is larger when the size of the training set is increasing. As more training samples are used for representation, the difference between each class becomes larger. Therefore, the teacher classifier is able to supervise the student more accurately.

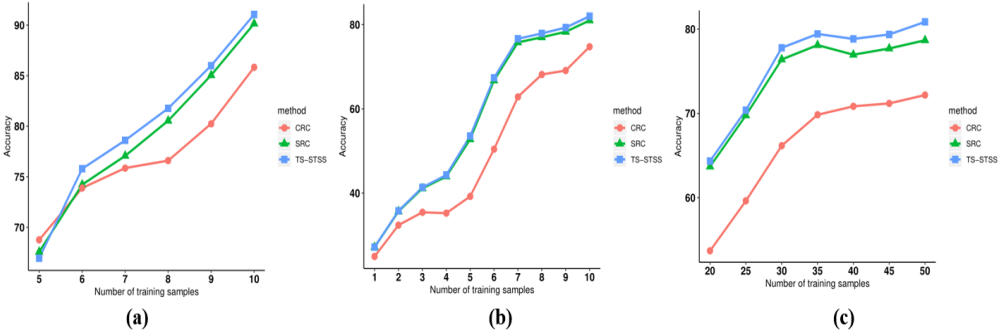

Figure 3: Recognition rate vs. increasing the number of training samples: (a) MUCT, (b) FEI, (c) COIL-100

## 3.4 Hand-writing Recognition

We applied the MNIST dataset to perform experiments on hand-writing recognition, where the results are displayed in Table 2. Here, the accuracy of SRC and CRC is 95.96% and 82.83% respectively. We can see that TS-STSS improves the recognition rate with the highest accuracy of 96.19%, which is higher than SRC and CRC by 0.23% and 13.36, respectively. Moreover, the proposed TS-STSS also achieves a better performance than TPSTR and STMS by 8.92% and 0.6%, correspondingly. Compared with other popular classifiers like K-SVD (82.87%), KNN (95.00%), and SVM (98.60%) TS-STSS is competitive as well.

## 3.5 Discussion

The experiments cover a variety of conditions and tasks in image classification, including face, object and hand-writing recognition, as well as different dataset sizes. The promising performances of our proposed method has been well proved. In addition, we can obtain the following inferences.

(1) Enlarging the training set is helpful in TS-STSS, since the implementation is based on sparse representation. As shown in Figure 3, the MUCT, FEI and COIL-100 datasets have no pre-split training and test sets, hence different training samples were utilized to train the classifiers. The accuracy keeps increasing as the training set becomes larger.

(2) The supervision of the teacher classifier is the key to improve the Two-Stage classification. As shown in Table 2, TS-STSS outperforms both TPSTR [5] and STMS [12] all but once (where for YouTubeFace the different with [12] is only 0.1%). This confirms our expectation that applying a consistent scoring criteria in two stages is beneficial to classification, while the conventional two-stage classifiers lack this.

(3) Compared to other linear methods, TS-STSS is very promising. Table 2 shows the recognition results of other popular linear classifiers, where TS-STSS consistently produces the highest accuracy in most cases. Furthermore, TS-STSS introduces only one additional parameter, the number of candidate class $k$, which can be set to an empirical value of $C/2$.

(4) Compared with SRC and CRC, TS-STSS outperforms both of them on all experiments, indicating that our proposed framework successfully integrates two weaker classifiers to form a stronger classifier in image classification.

## 4. Conclusion

In this paper, we propose a novel image classification framework named Two-Stage image classification Supervised by a Single-Teacher Single-Student model (TS-STSS). In the first stage, a candidate set of classes are chosen and the classification score vector is built using the L1-based SRC classifier (Teacher). Then, the L2-based CRC classifier (Student) represents the test sample using the candidate set in the second stage, under the supervision of the teacher classifier. In order to make a more precise score, we formulate it to the Single-Teacher Single-Student (STSS) problem. This image classification framework is able to combine two different weaker classifiers to form a stronger classifier. The experiments on five popular image datasets proved its effectiveness and promising capability on image classification, outperforming many other popular methods.

Currently, we have only implemented the proposed framework with linear sparse methods, SRC and CRC. It will be interesting to consider using other linear models as the teacher and student classifiers, i.e., dictionary learning [30], SVM, KNN, etc. Some nonlinear classifiers, like the ones based on deep neural networks, are potentially good choices as well. Despite our method utilizing image data, using deep features [20] ought to be helpful in real-world applications. We will continue to observe and explore these options in the future.

**Acknowledgements.** This work was supported by the University of Macau (MYRG2018-00053-FST).